\documentclass[11pt]{article}
\usepackage{coling2018}
\usepackage{times}
\usepackage{url}
\usepackage{latexsym}

\usepackage{amsmath}
\usepackage{amssymb}

\usepackage{graphicx}
\usepackage{subcaption}

\usepackage{amsfonts}

\usepackage{todonotes}

\usepackage[utf8]{inputenc}

\DeclareMathOperator*{\argmax}{argmax}

\usepackage{CJKutf8}

\slname{Turkish}

\title{Improving Named Entity Recognition by Jointly Learning to Disambiguate Morphological Tags}

\sltitle{Biçimbilimsel Etiketleri Ayrıştırmayı Birlikte Öğrenerek Varlık İsmi Tanıma Başarısını Artırmak}

\author{Onur Güngör \\
  Boğaziçi University, Istanbul, Turkey \\
  {\tt onurgu@boun.edu.tr} \\\AND
  Suzan Üsküdarlı \\
    Boğaziçi University, Istanbul, Turkey \\
  {\tt suzan.uskudarli@boun.edu.tr} \\\And
    Tunga Güngör \\
    Boğaziçi University, Istanbul, Turkey \\
  {\tt gungort@boun.edu.tr} \\}

\date{}

\begin{document}
\maketitle
\begin{abstract}
 % We improve the performance and scalability of a standard neural BiLSTM+CRF NER tagger for morphologically rich languages. 
  Previous studies have shown that linguistic features of a word such as possession, genitive or other grammatical cases can be employed in word representations of a 
  named entity recognition (NER) tagger
  %\textcolor{red}{composed of a bidirectional long short-term memory coupled with a conditional random field (Bi-LSTM+CRF)}\todo{bu cikabilir belki}  
  to improve the performance for morphologically rich languages. 
  However, these taggers require external morphological disambiguation (MD) tools to function which are hard to obtain or non-existent for many languages.
  % There is also the problem of portability as freely available morphological disambiguator software are mostly distributed for differing operating environments and hardware.
  In this work, we propose a model which alleviates the need for such disambiguators by jointly learning 
  NER and MD taggers in languages for which one can provide a list of candidate morphological analyses.
  We show that this can be done independent of the morphological annotation schemes, which differ among languages.
 Our experiments employing three different model architectures that join these two tasks show that joint learning improves NER performance.
 Furthermore, the morphological disambiguator's performance is shown to be competitive.
\end{abstract}

\makesltitle
\begin{slabstract}
Daha önceki çalışmalar, biçimbilimsel olarak zengin dillerdeki varlık ismi tanıma (VAT) başarısını artırmak için sözcüklerin iyelik, genitif ve benzeri hâllerinin kullanılabileceğini göstermiştir.
Ancak, bu türden varlık ismi tanıma işaretleyicilerinin çalışabilmesi için  elde edilmesi zor veya bazı diller için imkansız olan dışsal biçimbilimsel ayrıştırıcılara (BA) ihtiyaç vardır.
Bu çalışmada, bu tür ayrıştırıcılara olan ihtiyacı ortadan kaldırmak için VAT ve BA görevlerini aynı anda çözen ve aday biçimbilimsel çözümlemelerin sunulabildiği dillere uygulanabilen bir model önerilmektedir.
Bunun dillere göre değişen biçimbilimsel işaretleme şemalarından bağımsız olarak yapılabildiği gösterilmiştir.
Bu iki görevi aynı anda gerçekleştiren üç farklı model mimarisi kullanarak yaptığımız deneyler birlikte öğrenmenin VAT başarısını artırdığını göstermiştir.
Buna ek olarak, biçimbilimsel ayrıştırıcının başarısının önceki çalışmalarla karşılaştırılabilir olduğu görülmüştür.
\end{slabstract}

\section{Introduction}

\blfootnote{
    %
    % for review submission
    %
    %
    % % final paper: en-uk version 
    %
    % \hspace{-0.65cm}  % space normally used by the marker
    % This work is licenced under a Creative Commons 
    % Attribution 4.0 International Licence.
    % Licence details:
    % \url{http://creativecommons.org/licenses/by/4.0/}
    % 
    % % final paper: en-us version 
    %
    \hspace{-0.65cm}  % space normally used by the marker
    This work is licensed under a Creative Commons 
    Attribution 4.0 International License.
    License details:
    \url{http://creativecommons.org/licenses/by/4.0/}
}

Named entity recognition (NER) is the task of selecting the portions of text which refer to an entity that designate a person, location or organization.
This makes it a basic natural language processing (NLP) task closely related to relation extraction, knowledge base population, and entity linking.

Works that represent the current state of the art in NER generally start by representing words with pretrained word embeddings, embeddings which rely on surface form characters \cite{lample2016neural,ma2016end}. These architectures feed the word representations to a bidirectional long short-term memory layer (Bi-LSTM) to represent the context where the disambiguation between the possible entities is undertaken by decoding on trellis provided by a conditional random field (CRF) model.

When these models are trained and evaluated for morphologically rich languages (MRLs), it has been shown that using embeddings based on characters or linguistic properties of the word such as morphological features indicating a grammatical case improves the performance compared to using only pretrained word embeddings \cite{gungor2017morphological}.
Even though they provide a better approach for MRLs, they require an external morphological disambiguator for every language of interest, a requirement which can be hard or even impossible for some languages to satisfy. This is especially true for agglutinative languages where there can be many roots and morphological tag sequences for a single word.
Although there is an effort to provide a tool for POS tagging and lemmatization for many languages in a single format \cite{straka2017tokenizing}, it has been shown that there is a better approach for morphological tagging in terms of performance which can utilize the information in the context of the target word \cite{shen2016role}.

In this paper, we propose a model to jointly learn the NER and morphological disambiguation (MD) tasks to offer a solution to this problem. We design our model so that any language with a mechanism which can provide a number of candidate morphological analyses for a word can utilize our joint model.
This is easier compared to providing disambiguated morphological analyses because systems that disambiguate morphological analyses are harder to build.
Furthermore, we do not require the labels of each task to be present in the same dataset. One can easily train the part of the model which is responsible for the MD task in another -preferably larger- dataset and start with the pretrained model.
Our main contribution is to show that jointly disambiguating morphological tags and predicting the NER tags results in an equivalent level of performance compared to using externally provided morphological tags.

We give a survey of related work on the subject in Section \ref{sec:related_work}.
We explain our basic models and the proposed joint models in Section \ref{sec:model}. In Section \ref{sec:data}, we describe our dataset which is derived from a frequently used database in the literature. After running the experiments described in Section \ref{sec:results}, we observe that jointly training our model for NER and MD results in an increase in the NER performance. 

\section{Related Work}
\label{sec:related_work}

Early approaches to NER typically use several hand-crafted features such as capitalization, word length, gazetteer based features, and syntactic features (part-of-speech tags, chunk tags, etc.) \cite{mccallum2003early,finkel2005incorporating,humphreys1998university,appelt1995sri}. Some of them are data-driven approaches such as conditional random fields (CRF)~\cite {mccallum2003early,finkel2005incorporating}, maximum entropy~\cite{borthwick1999maximum}, bootstrapping~\cite{jiang2007instance,wu2009domain}, latent semantic association~\cite{guo2009domain}, and decision trees~\cite{szarvas2006multilingual}.

Recently, RNN based sequence taggers have dominated the state of the art in NER \cite{lample2016neural,ma2016end,huang2015bidirectional,Yang2016MultiTaskCS}. These approaches model the words as fixed length vectors and employ Bi-LSTM or GRU layers to obtain a characterization of the relevant context of the word to be labeled. These context vectors are then transferred to a CRF module after transforming into score vectors. However, in these studies, the morphological information present in the surface form of the word is handled only through the use of character based embeddings. Although this is not a limiting factor for languages which are not morphologically rich, it has been shown that employing morphologically disambiguated tags when representing words in a neural architecture improves the NER performance \cite{gungor2017morphological,strakova2016neural}.

There has been other approaches to the NER task for morphologically rich languages \cite{demir2014improving,seker-eryigit:2012:PAPERS,Yeniterzi:2011:EMT:2000976.2000995,tur2003statistical,hasan2009learning}. A study which can be considered as one of the first attempts in tackling NER for morphologically rich languages uses a hidden Markov model and takes the morphological tag sequence as input along with others like the surface form, capitalization features and similar features \cite{tur2003statistical}. In a study which basically depends on handcrafted features given to a CRF-based sequence tagger system, the word morphology was captured using the first and last three characters of the word as a feature resulting in an improvement in the NER tagging performance for Bengali \cite{hasan2009learning}. In another study \cite{Yeniterzi:2011:EMT:2000976.2000995}, a similar approach is taken with features generated using the output of an external morphological disambiguator and also shown to improve the performance. Another study \cite{seker-eryigit:2012:PAPERS} uses the same method but with a different approach for extracting morphological information, where they show an improvement over the previous state of the art results of \newcite{Yeniterzi:2011:EMT:2000976.2000995}. The first study focusing on morphologically rich languages to employ neural networks \cite{demir2014improving} contains a regularized averaged perceptron \cite{freund1999large} and relies on handcrafted rules along with pretrained word embeddings. However, they refrain from using output from external morphological disambiguators and only rely on the first and last few characters of a word as features. Our work in this paper differs from these studies as it does not rely on handcrafted features. We represent words as fixed length vectors, employ morphological information to disambiguate the correct morphological analysis, and then combine them in such a way to obtain a context vector to label with NER tags.

In a recent study on morphological disambiguation \cite{erayyildiz2016morphology}, the authors propose a two-layer network for prediction. 
In the first layer, they process the candidate morphological analyses along with the correctly predicted analyses of previous words and obtain a vector to be processed in the second layer. The second layer takes all vectors propagated from the previous words and computes a \texttt{softmax} function over positive and negative classes.
They predict the correct morphological analysis starting from the first word and use this prediction in the next word positions. The model is evaluated on a dataset manually labeled by the authors and considered as the state of the art for Turkish and competitive for French and German.
Our proposed model for morphological disambiguation relies on scoring the candidate morphological analyses to predict the correct one for a word in a sentence. We borrow this idea from \newcite{shen2016role}. In their study, they feed the word representations to a Bi-LSTM and obtain context embeddings for each position. Using these embeddings, they score each morphological analysis by calculating a similarity function reaching the state of the art or competitive results for Turkish, Russian and Arabic. 

Most of the work in morphological disambiguation or tagging strictly depend on their chosen specific output format for morphological analysis. This is due to the fragmented nature of computational approaches to morphological analysis for every language in the literature. 
However, we argue that our approach is immune to this problem as all of these output formats can be treated as a sequence. An example from Finnish is `\texttt{raha+\allowbreak [POS=NOUN]+\allowbreak [NUM=SG]+\allowbreak [CASE=ADE]}' \cite{silfverberg2016finnpos}, another from Turkish is `\texttt{Ankara+\allowbreak Noun+\allowbreak Prop+\allowbreak A3sg+\allowbreak Pnon+\allowbreak Loc}' \cite{oflazer1994two}, and one for Hungarian is `\texttt{hír+\allowbreak NOUN+\allowbreak Case=Nom+\allowbreak Number=Plur}' \cite{tron2005hunmorph}.
All of these can be split by the `+' symbol and transformed into a root and tag sequence.
Moreover, there is an attempt in the area to unify the morphological annotation along with syntax annotation across many languages which will contribute more towards a solution \cite{nivre2016universal}.

Many models targeting NLP tasks are designed to work independently although they usually employ linguistic information related with other tasks. Given that there are state of the art models which are similar in the sense that they all employ a sentence level Bi-LSTM, it is reasonable to hypothesize that jointly learning several tasks will improve the performance as shown in the literature \cite{Hashimoto2017AJM,luong2015multi}.
In a recent study, it has been suggested that using separate layers for separate tasks is better rather than using the same (or usually top) layer for all the tasks \cite{sogaard2016deep}.

\section{Models}
\label{sec:model}

We test our hypothesis by training a number of models where we choose to enable or disable the selected components and features\footnote{The code to replicate the experiment environment and the actual source code is published at \url{https://github.com/onurgu/joint-ner-and-md-tagger}}.
We start by explaining two basic models for each task: (i) a Bi-LSTM based sequence tagger where we predict the correct NER tags with a CRF (Section \ref{sec:ner_model}), (ii) a Bi-LSTM tagger which is used to represent the context for selecting the correct morphological analysis at the given position (Section \ref{sec:md_model}).
The joint models are combinations of these two basic models in various ways (Section \ref{sec:joint_models}).

\subsection{NER Model}
\label{sec:ner_model}

We formally define an input sentence as $X = (x_1, x_2, \dots, x_n)$ where each $x_i$ is a vector of size $l$ and the corresponding NER labels as $y_{\textsc{NER}} = (y_{\textsc{NER},1}, y_{\textsc{NER},2}, \dots, y_{\textsc{NER},n})$.
$x_i$ are then fed to a Bi-LSTM which is composed of two LSTMs \cite{Hochreiter1997} treating the input forwards and backwards. 
The output of this Bi-LSTM at position $i$, $h_i$, is a vector of size $2p$ where $p$ is the size of the LSTM cell. Further, we transform $h_i$ through a fully connected layer $\mathtt{FC}_{last}$ with $\mathtt{tanh}$ activations at the output to combine the left and right contexts into a vector of size $p$. This is followed by another fully connected layer to obtain a vector $s_i$ of size $K$, where $K$ is the number of the NER tags.

\begin{figure*}
    \centering

    \includegraphics[width=0.7\linewidth]{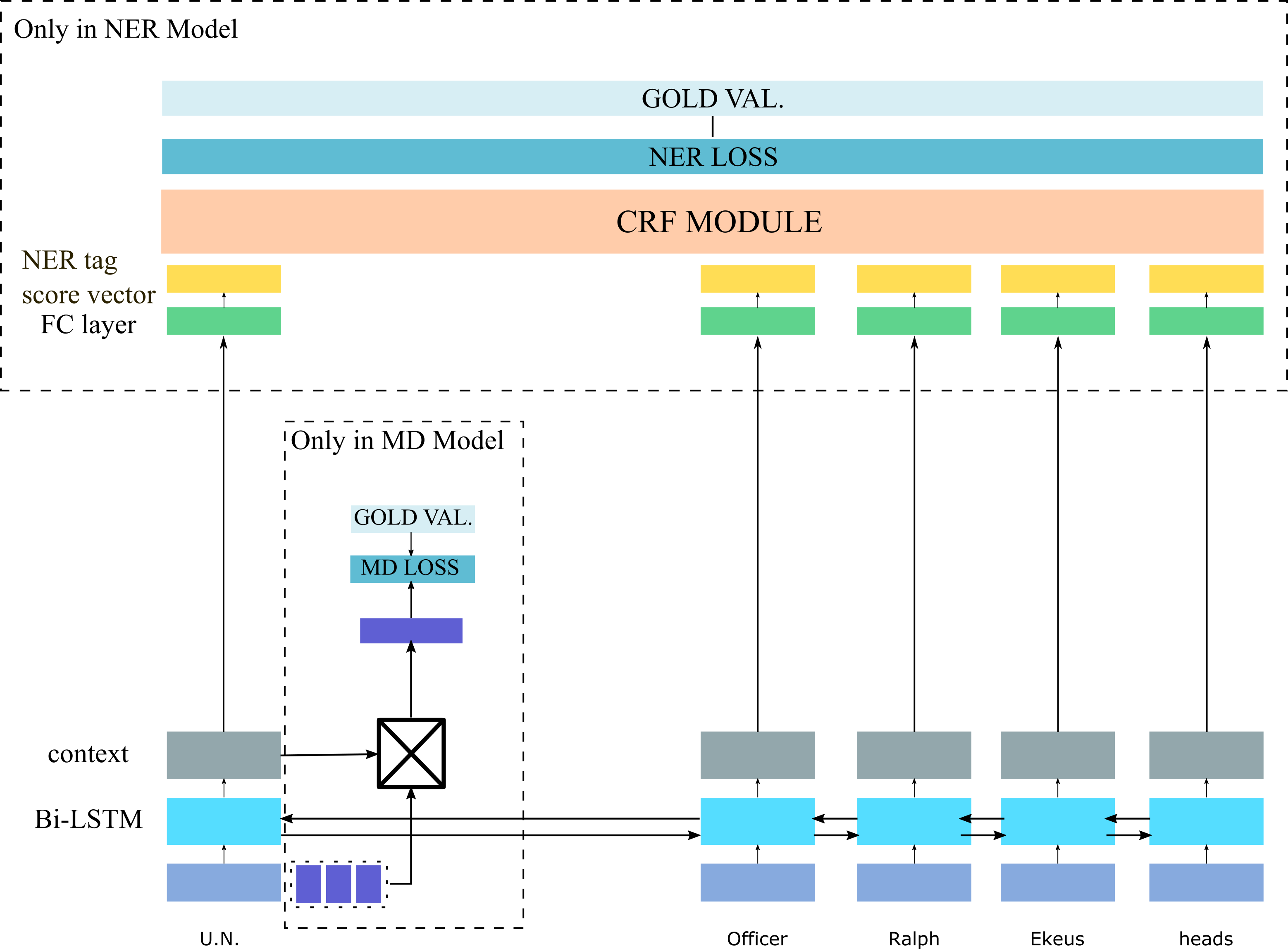}    
    \caption{Our basic models: \textsc{ner} and \textsc{md}. The portions of the model which are only active either for \textsc{ner} or \textsc{md} models are indicated with dashed lines. The symbol $\boxtimes$ represents the selection of $\mathtt{ma}_{ij^{*}}$.}\label{fig:basic_models}
\end{figure*}

We follow a conditional random field (CRF) based approach to model the dependencies between the consequent tokens \cite{lafferty2001conditional}. 
To do this, we take the vector $s_i$ at each position $i$ as the score vector of the corresponding word and aim to minimize the following loss function $\mathtt{loss}_{\textsc{NER}}(X, y_{\textsc{NER}})$ for a single sample sentence $X$:
\[
    -\sum_{i=0}^n A_{y_i,y_{i+1}} - \sum_{i=1}^n s_{i,y_i}  + log Z(X)
\]
where $A_{i,j}$ represents the score of a transition from tag $i$ to $j$, $Z(X) = \sum_{y^{'} \in \mathbb{Y}} \mathtt{exp} \left( \sum_{i=0}^n A_{y_i,y^{'}_{i+1}} + \sum_{i=1}^n s_{i,y^{'}_i} \right)$ where $\mathbb{Y}$ is the set of all possible label sequences. Using this model, we decode the most probable tagging sequence $y_{\textsc{NER}}^*$ as $\argmax_{\tilde{y}_{\textsc{NER}}} \mathtt{loss}_{\textsc{NER}}(X, \tilde{y}_{\textsc{NER}})$. We call this basic model the \textsc{ner} model \cite{lample2016neural} (see Figure \ref{fig:basic_models}).

In the remaining part of the section, we explain the details of the word representations used in this study.

\begin{figure}[h]
        \centering
        \includegraphics[width=0.5\linewidth]{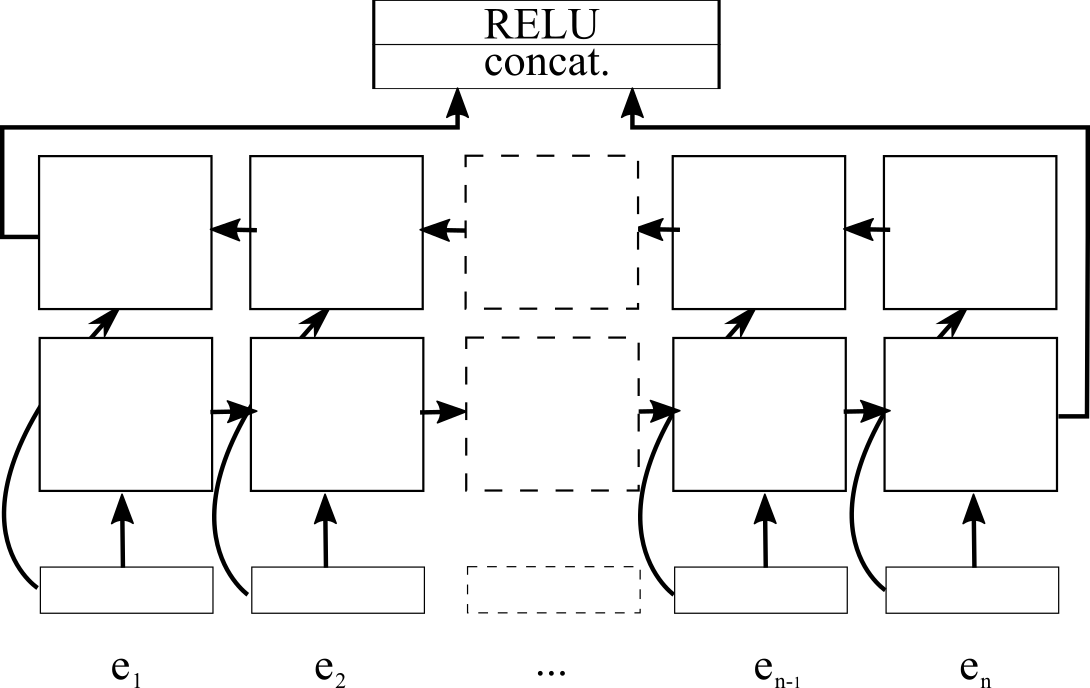}
        \caption{The basic model to generate representations for surface forms, roots, and morphological tag sequences. The input sequence $(e_1, e_2, \cdots, e_{n-1}, e_n)$ can either be the characters of the surface form, the characters of the root of the word, or the tags in the morphological tag sequence. RELU unit is active only for root and morphological tag sequences.\label{fig:unit_representations}}
\end{figure}

\textbf{Representing words.} As the default setting, we obtain word and character based embeddings as described below and combine them by concatenation.
For the first component, we allocate a word embedding vector of size $w_d$ for every word in our dataset. This can be loaded from a pretrained word embeddings database as is done frequently in the literature, but we chose to learn the word embeddings during training. 
The second component is generated from the surface forms. We feed the character sequence of the word into a Bi-LSTM as described at the beginning of this section. However, instead of using the outputs of LSTM cells at each position, we just take the last and the first cell outputs of the forward and backward LSTMs and concatenate them (Figure \ref{fig:unit_representations}). The resulting representation is two times the length of one character embedding length, $2\texttt{ch}_d$. This second component is in turn concatenated with the first component to obtain a word representation vector $x_i$ of size $w_d + 2 \texttt{ch}_d$. 

\textbf{External morphological features.} 
In order to compare our models with a previous method \cite{gungor2017morphological}, we utilize the golden morphological analysis provided with the dataset in addition to the word and character based embeddings and call this model \textsc{ext\_m\_feat}. The best approach reported by \newcite{gungor2017morphological} treats the string form of a morphological analysis as a sequence of characters and apply the process depicted in Figure \ref{fig:unit_representations}. 
For example, a morphological analysis in Hungarian is `\texttt{Magyar+PROPN+Case=Nom+Number=Sing}' in string form and can be split into a list of characters as \texttt{(M,a,g,y,a,r,+,P,R,O,P,N,+,C,a,s,e,=,N, o,m,+,N,u,m,b,e,r,=,S,i,n,g)}. 
Using the sequence of characters of the morphological analysis instead of the sequence of morphemes might seem counterintuitive at first glance.
However it has been argued that a benefit of treating morphological analyses as sequences of characters is the information conveyed by the characters within the tags. 
For example, in Turkish, the tags `\texttt{A3sg}' and `\texttt{A3pl}' indicate third person singular and third person plural where the leading two characters `\texttt{A3}' indicate third person agreement. 
This allows the model to represent the fragments of the tags which may improve the training performance. In this case, `\texttt{A3}' would represent the third person agreement independent of the singular or plural case.
The resulting vector representation is thus of length $2\texttt{mt}_d$  which is added to word and character based embeddings to obtain a word representation of $w_d + 2\texttt{ch}_d + 2\texttt{mt}_d $.

\subsection{MD Model}
\label{sec:md_model}

In this section, we describe our model for morphological disambiguation which is based on \cite{shen2016role}.
In this model, we are given a sentence $X$ in the same form as in the NER task, however we optimize the model to predict $y_{\textsc{MD}}$ where $y_{\textsc{MD},i}$ represents the correct morphological analysis out of the candidate morphological analyses for word $i$.
Like in the \textsc{ner} model, the \textsc{md} model also employs a Bi-LSTM layer to obtain context representations when fed with the word representations $x_i$ (Figure \ref{fig:basic_models}).
We define the candidate morphological analyses for word $i$ as $\texttt{ma}_i = \{ \texttt{ma}_{i,1}, \texttt{ma}_{i,2}, \cdots, \texttt{ma}_{i,j}, \cdots, \texttt{ma}_{i,K} \}$.
To determine the correct morphological analysis, we 
examine each morphological analysis output form to extract the root surface form and the morpheme sequence and generate the representation $\texttt{ma}_{ij}$ which we explain below.

We design this approach to be generalizable to many morphological analysis output forms described in Section \ref{sec:related_work}. We give an example from Turkish here: the unique analysis of the Turkish word ``Moda'da'' is ``\texttt{Moda+Noun+Prop+A3sg+Pnon+Loc}''. The word literally means `in Moda' (which is a district in Istanbul) and a common morpheme naming convention is used \cite{oflazer1994two}. So, we determine the root as `Moda' and the morpheme sequence as `\texttt{(Noun, Prop, A3sg, Pnon, Loc)}'. 
The root and the morpheme sequence are used to generate a representation as depicted in Figure \ref{fig:unit_representations}.
Except in this case the RELU activation function \cite{nair2010rectified} is also applied to the concatenation of the root and morpheme sequence representations. We choose the resulting representations $r_{ij}$ and $\texttt{ms}_{ij}$ to be of two times the length of a morpheme embedding $\texttt{mt}_d$. Furthermore, we add the root representation vector $r_{ij}$ and the morpheme sequence representation vector $\texttt{ms}_{ij}$ and apply hyperbolic tangent function (\texttt{tanh}), thus the morphological analysis representation $\texttt{ma}_{ij}$ is defined as follows $\mathtt{tanh}(r_{ij} + \texttt{ms}_{ij})$.

We then select the morphological analysis ${ma}_{ij^{*}}$ by performing a dot product with the context vector $h_i$: $\texttt{ma}_{ij^{*}} = \argmax_j h_i \cdot \texttt{ma}_{ij}$ when decoding. During training, the loss $\mathtt{loss}_{\textsc{MD}}(X, y_{\textsc{MD}})$ is calculated as 
\[
- \sum_{i=1}^n \mathtt{log}\,\mathtt{softmax}(\mathtt{mscore}_i)
\]
over  a score vector $\mathtt{mscore}_i$ such that $\mathtt{mscore}_{ij} = \{ h_i \cdot \texttt{ma}_{ij} \}$.

\subsection{Joint model for NER and MD}
\label{sec:joint_models}

We have experimented with three approaches for jointly learning NER and MD tasks. In this section, we explain the details of each approach.

\begin{figure*}[t]
    \centering
    \begin{subfigure}[t]{0.6\textwidth}
        \includegraphics[width=\textwidth]{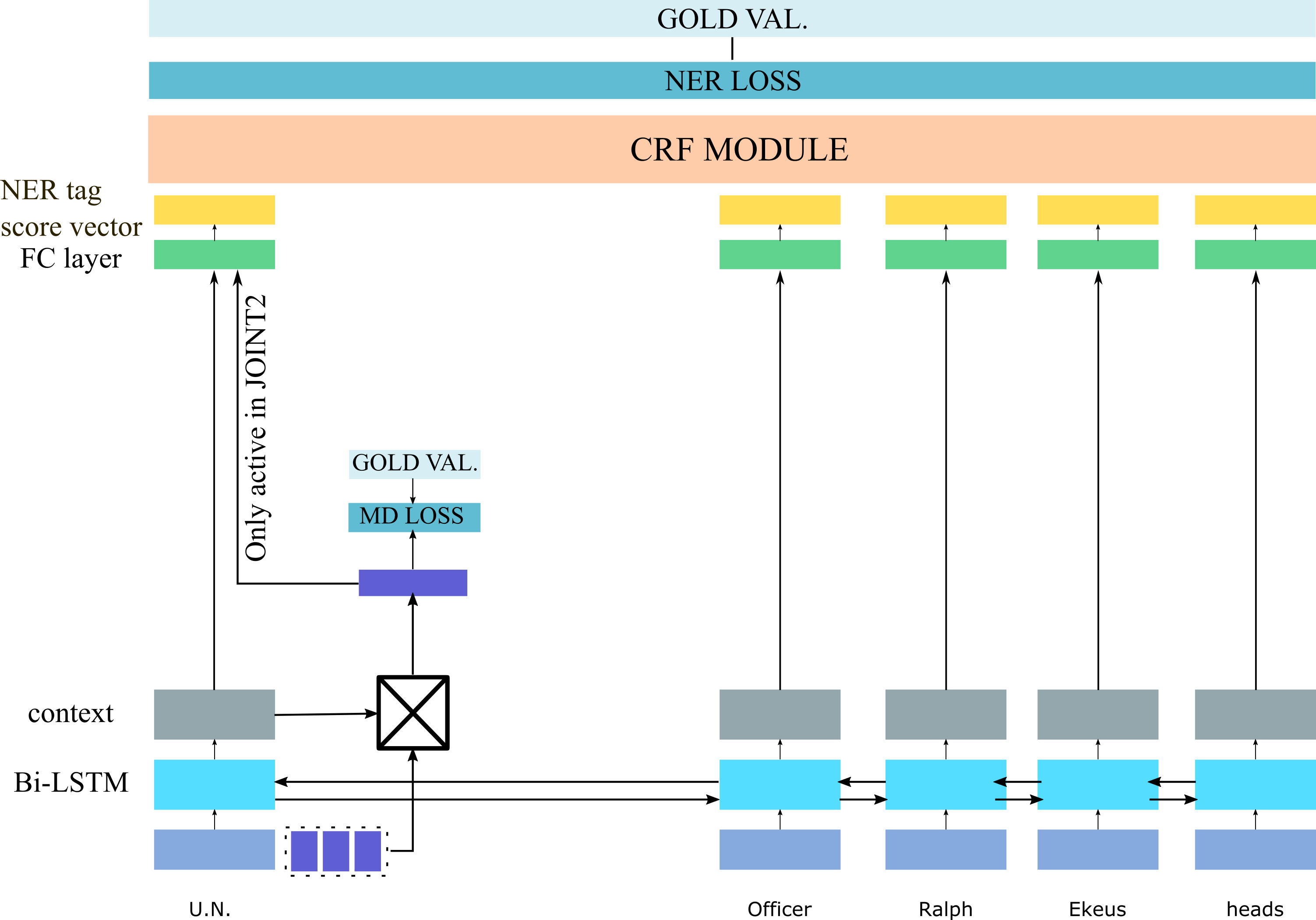}
        \caption{(i) Model \textsc{joint1}: two losses for two tasks sharing a Bi-LSTM. (ii) Model \textsc{joint2}: We concatenate the selected morphological analysis' vector representation to the last layer's context vector.}
        \label{fig:joint_models_joint1_joint2}
    \end{subfigure}
    ~ \par\bigskip %add desired spacing between images, e. g. ~, \quad, \qquad, \hfill etc. 
    \begin{subfigure}[t]{0.6\textwidth}
        \includegraphics[width=\textwidth]{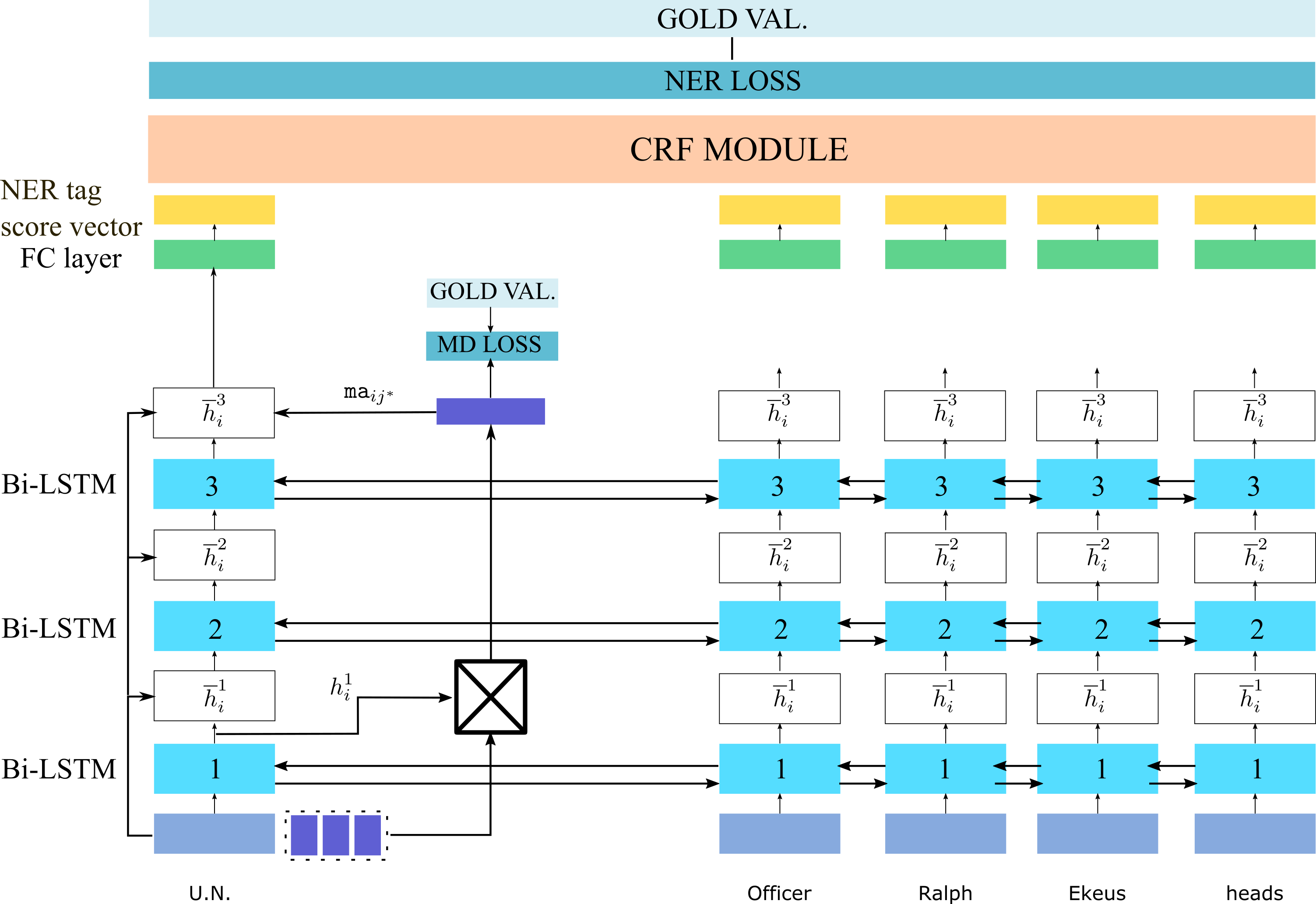}
        \caption{Model \textsc{j\_multi}: We employ shortcut connections and two more Bi-LSTM layers.}
        \label{fig:joint_models_multilayer}
    \end{subfigure}
    \caption{Our joint models: (a) \textsc{joint1} and \textsc{joint2} models (b) \textsc{j\_multi} model. The symbol $\boxtimes$ represents the selection of $\mathtt{ma}_{ij^{*}}$.}\label{fig:joint_models}
\end{figure*}

\textbf{Integration mode 1} - In this scheme, we employ a Bi-LSTM layer which is fed with word representations as in the basic models, \textsc{ner} and \textsc{md}. We then use the same context $h_i$ to calculate the losses separately for NER and MD as explained in Sections \ref{sec:ner_model} and \ref{sec:md_model}. We call this joint model \textsc{joint1} and show in Figure \ref{fig:joint_models_joint1_joint2}. 
We then learn the model parameters to optimize $\mathtt{loss}_{\textsc{JOINT1}}$
\[
\mathtt{loss}_{\textsc{NER}}(X, y_{\textsc{NER}})  + \mathtt{loss}_{\textsc{MD}}(X, y_{\textsc{MD}}). 
\]

\textbf{Integration mode 2} - As in the \textsc{joint1} model, this model also calculates separate losses for each task and sums them to obtain a single loss to optimize. However, we additionally concatenate the selected morphological analysis representation $\texttt{ma}_{i{j}^*}$ to $h_i$ before feeding it into the fully connected network with $\texttt{tanh}$ outputs as described in Section \ref{sec:ner_model}. The model is shown in Figure \ref{fig:joint_models_joint1_joint2}. The rationale of this concatenation is to facilitate information flow from the disambiguated morphological analysis. We call this model \textsc{joint2}. The loss function $\texttt{loss}_{\textsc{JOINT2}}$ of this model is then calculated similar to $\texttt{loss}_{\textsc{JOINT1}}$. 

\textbf{Multilayer and Shortcut Connections}. Our most complicated model is employing three Bi-LSTM layers instead of only one. We basically feed the output of the first layer $h_i^1$ to layer 2, the output of the second layer $h_i^2$ to layer 3. In addition to this, we transfer the word representation $x_i$ to all layer inputs and concatenate with $h_i^{level}$ to obtain $\overline{h}_i^{level}$. When processing to obtain the third layer's output $\overline{h}_i^3$, we also concatenate the selected morphological analysis representation $\texttt{ma}_{i{j}^*}$ to $h_i^3$ in addition to $x_i$. This is done to propagate the information gained from the disambiguated morphological analysis to the last layer of the network. We use the first layer's output $h_i^1$ when calculating $\texttt{mscore}_i$ as shown to be better for a variety of tasks \cite{Hashimoto2017AJM}.  We call this model \textsc{j\_multi} and depict in Figure \ref{fig:joint_models_multilayer}.

\section{Data}
\label{sec:data}

To test our proposed model, we derived a new dataset based on a dataset commonly used in the literature for the NER task for Turkish \cite{tur2003statistical}. This dataset contains sentences from the online edition of a Turkish national newspaper with NER labels. The creators of the dataset also provide a golden morphological analysis along with each word. However, golden morphological analyses in this dataset are sometimes erroneous. For example, words which are inflections of foreign words are usually problematic. An example is ``Hillary'nin'' which is the genitive case for the word ``Hillary''. It has been incorrectly labeled as if it is in nominal case. 
Also, when the surface form is a number in some noun case, like ``98'e'' which is the dative case for the number ninety eight, the morphological analysis is almost always nominal. We believe the reason for this is the incorrect handling of the quote character when preparing the original version.

In our study, we have first divided the training portion of the original dataset into training and development sets. We then augmented these portions using candidate morphological analyses for each word with a commonly used morphological analyzer \cite{oflazer1994two}. Unfortunately, the golden morphological analyses in about 5\% of the word tokens were not found in these candidate analyses. To mitigate this issue, we listed the most frequent contexts where a specific mismatch happens, selected the most suitable morphological analysis out of the candidates for each context, thus providing a solution to the mismatch. We then automatically corrected all contexts with a mismatch which has a solution in our solution database. Although we tried to give the utmost attention to selecting the best solution, some of our solutions might be problematic. Thus, we share the data, the scripts and the tool which helps the user to select a solution as described for academic use and examination\footnote{The data can be found at \url{https://github.com/onurgu/joint-ner-and-md-tagger}}.
Unfortunately, there were still left a few hundred mismatches. As providing a solution for them required a lot of manual work and would only save 1-2 sentences for each, we just removed any sentence that contains any of these mismatches. 
This way, we have retained 25511 out of 28835 sentences in the original dataset for training, 2953 of 3336 for development and 2913 of 3328 for test. By this process, despite losing some of the sentences, we have built a new dataset with both the NER labels and the candidate morphological analyses which have correct golden labels.

\subsection{Training}

We implemented the model using the DyNet Neural Network Toolkit in Python. The model parameters are basically the word embeddings, the parameters of Bi-LSTMs, the weights of the fully connected layer $\mathtt{FC}_{last}$, and the CRF transition matrix $A$. We trained by calculating the gradients of the loss for a batch of five sentences consisting of surface forms and its associated NER and/or MD labels
and updated the parameters with Adam \cite{adam_kingma2014} for 50 epochs and reported the performance on test set of the model with the highest development set performance. We applied dropout \cite{dropout_srivastava2014} with probability 0.5 to the word representations $x_i$. To facilitate the reproducibility of our work, we also provide our system as a virtual environment\footnote{You can obtain our implementation and find more information about how to use our virtual environment at \url{https://github.com/onurgu/joint-ner-and-md-tagger}.} that provides the same environment on which we evaluated our system in an open manner.

\section{Results}
\label{sec:results}

To test our approach, we train and evaluate every model for 10 times and report the mean F1-measure value for named entity recognition and accuracy for morphological disambiguation. This is done to decrease the potential negative effects of random initialization of model parameters as shown in the literature \cite{reimers2017optimal}. To accomplish this given our limited computing resources, we set the parameter dimension sizes to 10 and do not employ pretrained word embeddings.

The results are shown in Table \ref{tab:results_NER}. We see that the mean NER performance increases in joint models. 
We see that the \textsc{joint2} model is performing better than just calculating two losses at the last layer as we did in the \textsc{joint1} model. However, applying the Welch's t-test between the \textsc{joint1} and \textsc{joint2} runs does not strongly imply this difference $(\textit{p} = .24)$. 
Adding multiple Bi-LSTM layers to \textsc{joint2} and obtaining \textsc{j\_multi} also helped and achieved the best score among our joint models\footnote{One can wonder whether this performance improvement could be due to an increase in the total number of parameters of the model. We saw that the increase is negligible as it only accounted for a 2\% increase.}. Employing Welch's t-test confirms the significance of this difference with other joint models, $p < .05$ for each pair.

\begin{table}[h]
\centering
\begin{tabular}{|p{2cm}|p{2cm}|}

                        \hline
   \multicolumn{2}{|l|}{This work} \\
                        \hline
   Model      & \multicolumn{1}{c|}{Mean F1-measure} \\
                        \hline
\textsc{ner} & \multicolumn{1}{c|}{81.07} \\
                        \hline
\textsc{joint1} & \multicolumn{1}{c|}{81.28}  \\
                        \hline
\textsc{joint2} & \multicolumn{1}{c|}{81.84}  \\
                        \hline
\textsc{j\_multi}  &       \multicolumn{1}{c|}{\textbf{83.21}}  \\
                        \hline
   \multicolumn{2}{|l|}{Previous work} \\
                        \hline
\textsc{ext\_m\_feat} &    \multicolumn{1}{c|}{\textbf{83.47}}    \\
                        \hline
\end{tabular}
\caption{Evaluation of our models for NER performance with our dataset.
We report F1-measure results over the test portion of our dataset averaged over 10 replications of the training with the same hyper parameters.\label{tab:results_NER}}
\end{table}

To make a comparison with a previous method \cite{gungor2017morphological}, we also evaluated a model where the golden morphological analysis in the corpus is represented as a vector and included in the word representation $x_i$, namely \textsc{ext\_m\_feat} (see Section \ref{sec:ner_model}). As one can see from the table, it achieved the best results compared to our joint models. However, we cannot confirm the difference between \textsc{ext\_m\_feat} and \textsc{j\_multi} models as the calculated $p$ is well above $.05$.
Thus our best performing model \textsc{j\_multi} is performing at a competitive level with an additional advantage of disambiguating the morphological tags while predicting the NER tags. This also serves as another confirmation to the hypothesis that employing linguistic information such as morphological features leads to an increase in the NER performance.

\begin{table}[h]
\centering
\begin{tabular}{|p{4cm}|p{1.5cm}|}

                        \hline
                        \multicolumn{2}{|l|}{This work} \\
			\hline
            \multicolumn{1}{|l|}{Model} & \multicolumn{1}{|p{1.5cm}|}{Mean Accuracy}  \\
                        \hline
\textsc{md}                       &    88.61       \\
                        \hline
\textsc{joint1}               &     88.17         \\
                        \hline
\textsc{joint2}                 &       86.86      \\
                        \hline
\textsc{j\_multi}          &   88.05   \\
                        \hline
                        \multicolumn{2}{|l|}{Previous work} \\
                        \hline
\newcite{yuret2006decision_lists}  &   89.55       \\
                        \hline
\newcite{shen2016role}     &   \textbf{91.03}   \\
\hline
\end{tabular}
\caption{Evaluation of our models for MD performance. 
As in the NER evaluation, we report accuracies over the test dataset averaged over 10 replications of the training.\label{tab:results_MD}}
\end{table}

To evaluate the performance of morphological disambiguation, we have tested the MD performance of our models, which are trained with the training portion of our dataset, on the test portion of a frequently used dataset \cite{yuret2006decision_lists}.
As can be seen from Table \ref{tab:results_MD}, we are very close to the state of the art MD performance even if we only trained with a low number of parameters as stated in the beginning of this section. We have to also note that in contrast with the NER task, the MD task did not enjoy a performance increase from joint learning. 
The mean morphological disambiguation accuracies for the test portion of our dataset also suggests the same, all hovering around 77\% without much change.
This might be due to the fact that NER can utilize the disambiguated morphological analysis of a word to predict the correct label, however a correctly predicted NER label does not contribute to the disambiguation of the word's morphology.

\section{Conclusions}

In this work, we propose a joint model of NER and MD tasks that removes the need for external morphological disambiguators. The method is applicable to every language given that one can provide the candidate morphological analyses for a word, making this approach portable to many languages. We have also shown that joint learning leads to an increase in the NER tagging performance.
However, there is more work to do as we are still bound to language specific tools in obtaining the list of candidate morphological analyses. Generating the list of candidate analyses within the model, testing our hypothesis on other morphologically rich languages, and testing with models which have higher number of parameters are left for future work.

\section*{Acknowledgements}

This study was supported by the Boğaziçi University Research Fund (BAP 13083).

\bibliographystyle{acl}
\bibliography{coling2018}

\end{document}